\documentclass[runningheads]{llncs}

\usepackage{eccv}

\usepackage{eccvabbrv}

\usepackage{graphicx}
\usepackage{booktabs}
\usepackage{float}
\usepackage[accsupp]{axessibility}  

\usepackage[pagebackref,breaklinks,colorlinks,citecolor=eccvblue]{hyperref}

\usepackage{orcidlink}

\begin{document}

\title{Controllable Talking Face Generation by Implicit Facial Keypoints Editing} 

\titlerunning{Abbreviated paper title}

\author{Dong Zhao\inst{1} \and
Jiaying Shi\inst{1} \and
Wenjun Li\inst{1} \and
Shudong Wang \and
Shenghui Xu\inst{1} \and
Zhaoming Pan\inst{1}
}

\authorrunning{F.~Author et al.}

\institute{NetEase Media Technology (Beijing) Co., Ltd.
\email{\{zhaodong03,shijiaying,liwenjun01,xushenghui,panzhaoming\}@corp.netease.com}}

\maketitle

\begin{abstract}
  Audio-driven talking face generation has garnered significant interest within the domain of digital human research. Existing methods are encumbered by intricate model architectures that are intricately dependent on each other, complicating the process of re-editing image or video inputs. In this work, we present ControlTalk, a talking face generation method to control face expression deformation based on driven audio, which can construct the head pose and facial expression including lip motion for both single image or sequential video inputs in a unified manner. By utilizing a pre-trained video synthesis renderer and proposing the lightweight adaptation, ControlTalk achieves precise and naturalistic lip synchronization while enabling quantitative control over mouth opening shape. Our experiments show that our method is superior to state-of-the-art performance on widely used benchmarks, including HDTF and MEAD. The parameterized adaptation demonstrates remarkable generalization capabilities, effectively handling expression deformation across same-ID and cross-ID scenarios, and extending its utility to out-of-domain portraits, regardless of languages. Code is available at \href{https://github.com/NetEase-Media/ControlTalk}{ControlTalk}. 
\end{abstract}

\section{Introduction}

Recently, video generation with artificial intelligence(AI) has been attracting increasing attention and its applications are also expanding in various fields~\cite{blattmann2023stable,lu2023vdt}. In particular, audio-driven talking face generation, such as visual dubbing~\cite{prajwal2020lip,zhang2023dinet,cheng2022videoretalking}, and human animation~\cite{guo2021adnerf,ye2023geneface}, is highly promising and able to provide convenience to human life in the fields of education, news and media~\cite{zhou2020makelttalk,zhang2023sadtalker}. Audio-driven talking face generation aims to produce synchronized speaking videos. Though great progress has been made in generating natural face motion, most previous methods are typically complicated due to multiple processing stages with prolonged training times and extensive computational resources~\cite{zhang2023sadtalker,song2022everybodys}. 

For both video dubbing and single image-based talking face generation, it is very challenging to smoothly control head poses while generating lip-synced videos in a unified manner. Previous single image-based talking-head generation methods~\cite{ji2022eamm,zhou2021pcavs,ma2023dreamtalk} are focused on audio-visual synchronization based on a pose reference sequence, while other recent works such as SadTalker~\cite{zhang2023sadtalker} generate pose parameters in a learnable way. 
Furthermore, talking face generation methods~\cite{zhang2023dinet,cheng2022videoretalking,ye2023geneface} that rely on video clips to maintain original poses only learn individual lip motions, which is not applicable without character's video clips.
Additionally, these methods require multiple steps in the training and finetuning stage, which makes the generated results vulnerable to accumulated errors. 
The talking face performance could introduce errors at every stage, amplifying the inaccuracy of the effect and thereby compromising the fidelity of the final output~\cite{guo2021adnerf,wang2021audio2head,song2022everybodys,gururani2023space,zhang2023sadtalker,ma2023dreamtalk}.

To address the above limitations, a desirable approach should efficiently and flexibly combine single image-based and video-based inputs for talking face generation as illustrated in Fig.~\ref{fig:framework}. We also propose a lightweight parameterized module to simplify the generation process.  There are three key advantages.
Firstly, the lightweight adaptation is not sensitive to image resolution and is used to predict implicit facial keypoints. Therefore, we can readily apply ControlTalk to any scale of image resolution by modifying the input of pre-trained models. 
Secondly, parameterized adaptation allows for flexible control of mouth shape which could be more suitable for different speakers. To the best of our knowledge, our study is the first to control different mouth-opening shapes for the same phonemes.
Lastly, obtaining the pre-trained models is simpler and more adaptable in unknown scenarios, such as other languages and out-of-domain images that are not real humans without training.

We propose a lip synchronization method ControlTalk to unify both single image and video-based talking face generation, which involves two kinds of pre-trained models. The first is audio encoder~\cite{hubertfeature} for input speech feature extraction. The other is a video synthesis renderer face-vid2vid~\cite{wang2021facevid2vid} for face motion extraction and parameterized face renderer. As shown in Fig.~\ref{fig:framework}, we propose a learnable \textit{Audio2Exp} module as a lightweight adaptation to map audio and original face expression to the enhanced expression points, which could be rendered to talking face images with other 3D implicit points including head pose, etc. Our approach is trained using speaking videos but can be quickly transferred to the single image-based task by replacing 3D implicit points. The main contributions and innovations of our work are as follows:

\begin{figure}[tb]
\centering
\includegraphics[height=6.7cm]{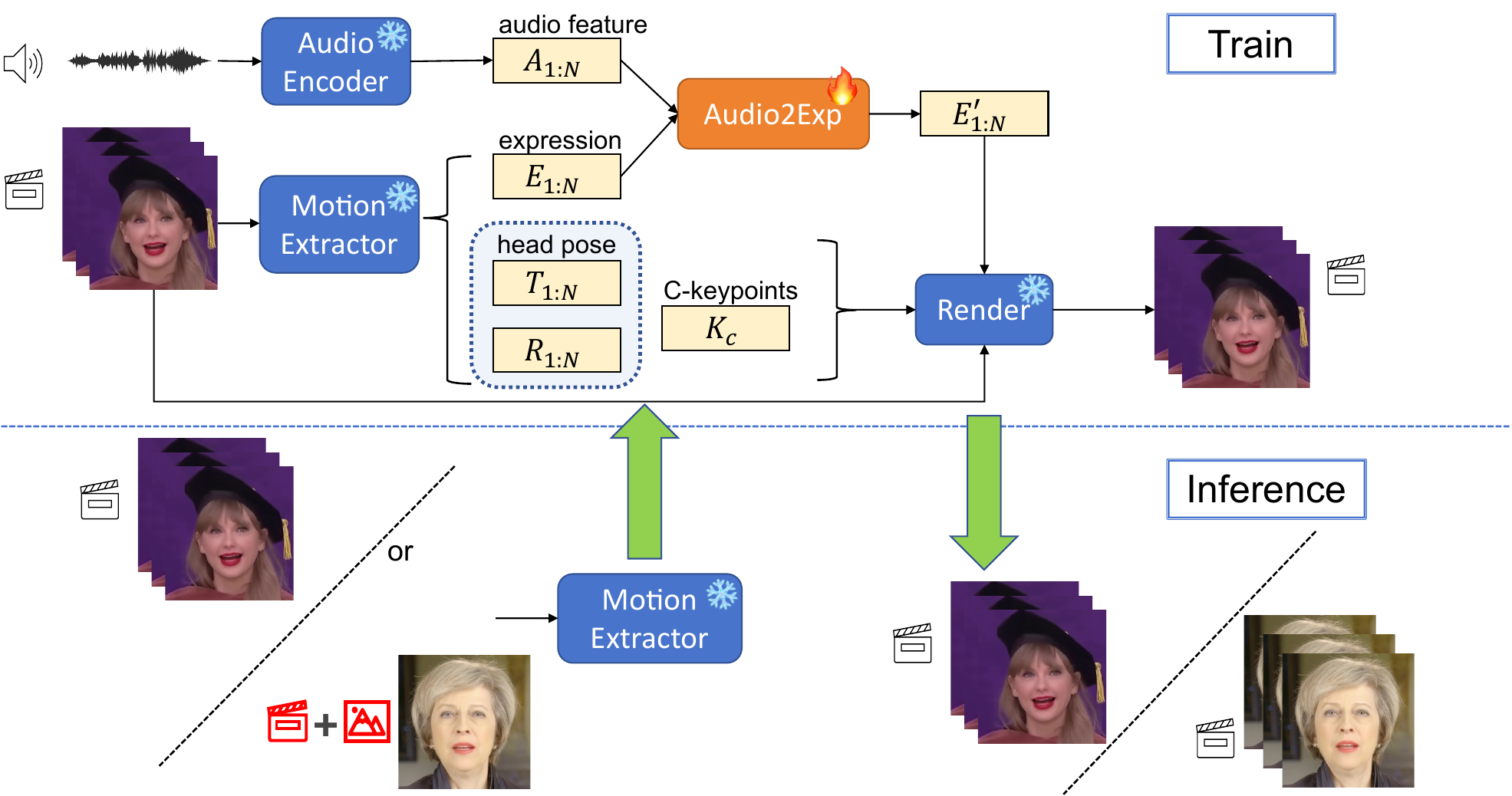}
\caption{\textbf{An overview of ControlTalk.} Our method consists of 4 modules, but only \textit{Audio2Exp} participates in training to simplify the whole process. In the training process, audio and video are used as inputs, and the speech features and parameterized coefficients are extracted by the pre-trained model respectively, which are subsequently converted into lip-synced expression coefficients through \textit{Audio2Exp}. Finally, the input video frame and parameterized coefficients including new expression coefficients would be rendered to the generated talking face video. In the inference phase, 
image input is also supported with driven motions.}
\label{fig:framework}
\end{figure}

 - We have proposed a new lip synchronization method ControlTalk that edits parameterized facial keypoints to achieve efficient talking face generation. Our method simplifies the generation process with lightweight adaptation, allowing more flexible control of mouth shape and reducing the possibility of accumulated errors.

 - Compared to current methods, our approach offers greater versatility and adaptability, as it accommodates input from both images and videos. This generalization capability allows for a broader range of potential applications across various scenarios and requirements.

 - Experiments have proven that our ControlTalk outperforms previous methods in terms of both lip synchronization and video quality, which can be extended to high-resolution video, and can be applied to multiple characters and languages.

\section{Related Work}
\textbf{Audio-driven Single Image-based Talking Face Generation. }In the field of audio-driven single-image talking face video generation, several researchers have made notable contributions.~\cite{chen2019hierarchical} advanced speaking face generation by employing a cascaded structure and attention mechanism to address the limitations of previous methods. MakeltTalk~\cite{zhou2020makelttalk} successfully separated speaking content from speaker identity and utilized facial landmarks as an intermediate representation to generate more realistic and natural speaker-aware facial expressions and head pose animations.~\cite{hdtf} proposed a novel flow-guided framework based on 3D Morphable Model (3DMM)~\cite{blanz2023morphable}, utilizing a new large-scale high-definition dataset to synthesize high-quality, high-definition one-shot talking face videos.  Audio2head~\cite{wang2021audio2head} improved visual quality and head movement realism by introducing a keypoint motion field representation. Recently,~\cite{gururani2023space} used LSTM to predict normalized facial feature point movements, converting them into the implicit keypoints of a facial animation model to generate facial animation videos. SadTalker~\cite{zhang2023sadtalker} introduced a motion coefficient representation method based on 3DMM and developed ExpNet and PoseVAE to generate realistic motion coefficients from audio, along with a 3D-aware facial renderer~\cite{wang2021facevid2vid} for high-quality single-image talking head video generation. DreamTalk~\cite{ma2023dreamtalk} is an expression-speaking avatar generation framework based on a diffusion probability model, leveraging the model to deliver high performance across various speaking styles while reducing reliance on costly style references. These approaches have significantly advanced the field of audio-driven single-person video generation, providing valuable insights and substantial progress. Although there are some previous works~\cite{hdtf,ren2021pirenderer} utilize 3DMM as the implicit representation, their method still faces the problem of inaccurate expressions with high-dimensional coefficients.
\\\noindent\textbf{Audio-driven Video-based Talking Face Generation. } The task of generating a talking face aims to synthesize facial video according to speech audio. Early efforts by Taylor et al.~\cite{taylor2017deep} explored the conversion of audio sequences into phoneme sequences to create adaptable talking avatars capable of speaking multiple languages. Videoretalking, LipGAN, and Wav2Lip~\cite{cheng2022videoretalking,kr2019towards,prajwal2020lip} mainly focus on producing an accurate mimic of the lip movements of any individual in a dynamic speaking face video by leveraging a lip-sync discriminator. To render more high-fidelity faces, DINet~\cite{zhang2023dinet} introduced a Deformation Inpainting Network that enables visually realistic dubbing on high-resolution videos. However, one drawback is that if the mouth area overlaps with the background, artifacts may be generated on the outside of the face. More recently, DiffTalk~\cite{shen2023difftalk} has employed implicit diffusion models to achieve high visual quality, but at the cost of compromised lip-sync, particularly when generating faces across different generations. In order to achieve a more realistic synthesis, FACIAL~\cite{zhang2021facial} and ~\cite{song2022everybodys} utilized audio to regress parameters in 3D face models. However, there are still challenges to be addressed to achieve both realistic expression and accurate lip movement in the generated videos. To enhance video quality, ADNeRF~\cite{guo2021adnerf} RAD-NeRF~\cite{tang2022radnerf} and Geneface~\cite{ye2023geneface} improved video quality by employing an audio-driven neural radiance fields (NeRF) model to generate high-quality talking-head videos based on audio input.
In our work, we introduce a lightweight adaptation module to achieve efficient and effective lip synchronization for both image and video as input.

\section{Method}

ControlTalk is a lip synchronization method that edits implicit facial keypoints to achieve efficient talking face generation, which simplifies the generation process with lightweight adaptation while preserving the generated image quality of awesome renderer~\cite{wang2021facevid2vid}.
In this section, we first introduce the basic structure of ControlTalk in Sec.~\ref{sec:ControlTalk} and then describe how we apply a lightweight {\it Audio2Exp} network to locally change expression coefficients in Section~\ref{sec:Audio2exp}. 
Moreover, it allows for nuanced control over the open scale of talking mouth by the adjustable parameters, facilitating a more consistent and realistic representation, which is detailed in Section~\ref{sec:mouth_scale}.

\subsection{ControlTalk}
\label{sec:ControlTalk}
3D information is essential for enhancing the realism of generated videos since the real talking face videos are captured in the 3D environment. Previous works like~\cite{zhang2023sadtalker} have considered the space of the predicted 3DMM as the intermediate representation. Nevertheless, there is also a need for a mapping network that transfers to the implicit features, which may accumulate errors. Inspired by this, we consider an unsupervised keypoint representation~\cite{wang2021facevid2vid} to directly render the face shown in Fig.~\ref{fig:framework}.

Our proposed method generates the talking face inheriting the motion of  the input video, meanwhile, we also pay attention to the audio for deforming lip expression into a neutral appearance. Particularly, benefiting from the parameterized design, our method can be flexibly adapted to image input with driving audio and video motions. 
Let $\left \{ d_{1}, d_{2},..., d_{N} \right \} $ be the input video, where $d_{i}$ is the each frame, and $N$ is the total frame number. Let $\left \{ a_{1}, a_{2},..., a_{N} \right \} $ be the driven audio, which has been aligned with driven video. 
Our goal is to generate an output video $\left \{ y_{1}, y_{2},..., y_{N} \right \} $, where the identity and the motions in  $y_{i}$ is inherited from $d_{i}$. Especially, in the mouth region, the lip motion is synthesized based on driven audio.

Firstly, we encode the audio's speech feature $A$ and extract the main parts of facial motions including expressions $E$ and other geometric coefficients of a person, such as head pose, and canonical keypoints(C-keypoints). Secondly, we apply {\it Audio2Exp} network to predict lip expressions $E'$ based on input speech feature $A$ and original expressions $E$. Finally, the combined keypoints would re-edit the input image and render a new talking face with geometric coefficients jointly.

\subsection{Audio2Exp}
\label{sec:Audio2exp}

Synthesizing a talking face video requires identifying the specific person, such as face appearance, pose, and expression. As shown in Fig.~\ref{fig:framework}, in the training stage, the input video and audio are aligned and we can extract the 3D facial motions based on pre-trained motion extractor~\cite{wang2021facevid2vid}. Given a frame $d_{i}$, the 3D motions $K_{i}$ represent pose and expression, which are composed of four components: expression deformation $E_{i}$, translation $T_{i}$,  rotation matrix $R_{i}$, and identity-specific canonical keypoints $K_{c}$. These components are then combined as follows:

\begin{equation}
    \label{eq:facevid2vid}
    K_{i} =R_{i} K_{c}+T_{i}+E_{i}.
\end{equation}
Our goal is to use the above 3D motions $K_{i}$ to render the input video frame into a lip-synced video, which is defined as:

\begin{equation}
    \label{eq:render}
    y = f_{r}(K, d)=f_{r}(RK_{c}+T+E', d),
\end{equation}
where $f_{r}$ represents the face renderer, and $E'$ is the lip-synced face expressions.
Given the original expression deformation $E_{i}$,  our {\it Audio2Exp} network extracts the motion-related information based on the input audio to predict the new expression $E_{i}'$.

We have observed that even small changes in $E'$ can have a great impact on the generated face images based on Eq.~\ref{eq:render}, such as distortion of facial appearance, etc. Therefore, the {\it Audio2Exp} is designed to predict a bias $ \Delta E_{i}$ of expression deformation through a progressive method.

\begin{equation}
    \label{eq:audio2exp}
    E_{i}' = E_{i} + \alpha\cdot  \Delta E_{i},
\end{equation}
where $\alpha$ would gradually increase from 0 to 1 as the network training. 
In the meantime, {\it Audio2Exp} is implemented through {\it Zero Module} that is a unique type of {\it Linear} layer that progressively grows parameters from zero to optimized values in a learnable way~\cite{controlnet}. This special training strategy guarantees that slight changes in $E_{i}'$ are not incorporated into the deep features at the start of training, while also making no effect in the downstream stage to render a talking face.

\subsection{Adjustable Talking Mouth}
\label{sec:mouth_scale}
We observe that parameterized adaptation allows for flexible control of mouth shape.
As shown in Eq.~\ref{eq:audio2exp}, the $\alpha$ for expression deformation is changed during the training process.  This design stabilizes the training phase, maintaining predictability and control over the impact of variables on the overall model. 
Therefore, it is intuitive that we can change the value of $\alpha$ to control the impact of audio on the original expression coefficient $E$. Based on this idea, it offers a more flexible way to regulate the size of the talking mouth.

\begin{figure}[H]
\centering
\includegraphics[height=5cm]{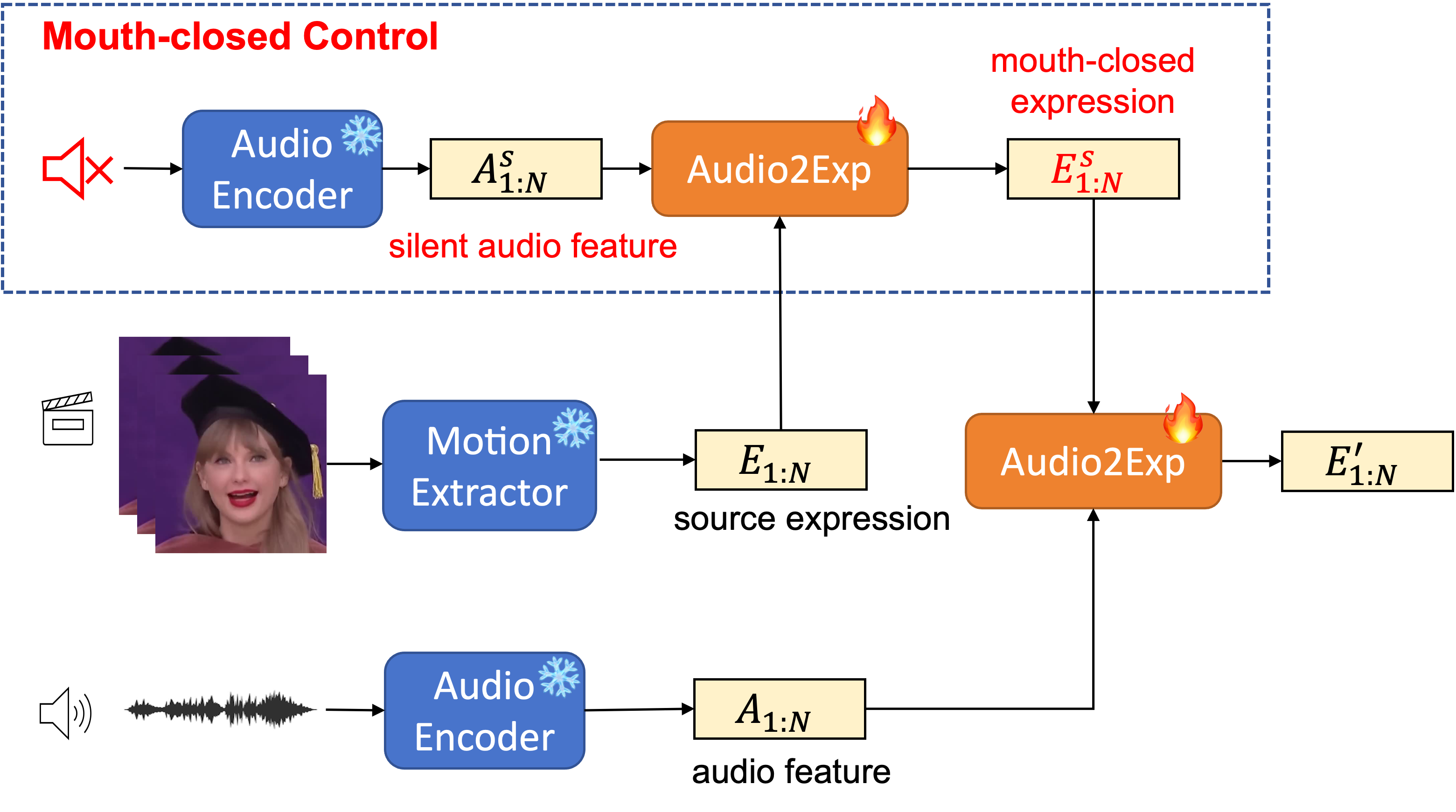}
\caption{\textbf{Silent audio training for adjustable talking mouth.} Silent audio would first control the predicted expression, and then the final expression is synchronized by input audio through {\it Audio2Exp}.  }
\label{fig:mute_audio}
\end{figure}

In addition, we have found that in facial image rendering, for the whole face area, although the expression coefficients for the full face region are not linearly separable, the audio-related components significantly influence the changes in mouth shape.
Moreover, by adjusting the number of the bias coefficient $\alpha$ , the degree of mouth opening is correspondingly affected. The comparison cases are detailed in Sec.~\ref{sec:exp_mouth_amplitude}.

To get better control of mouth shape, we also take advantage of the silent audio. Because different speakers have different speaking habits within the training dataset, there can be significant variations in the mouth shapes corresponding to the same phonemes. 
However, we aim to control the size of the mouth shape by the bias coefficient $\alpha$, so it is crucial to transfer all training videos within the same distribution. Consequently, our model is designed with a dual training approach under the guidance of silent audio as shown in Fig.~\ref{fig:mute_audio}. To our surprise, this method also ensures that our model can handle lip motion under silent audio effectively, resulting in more stable performance.

\subsection{Losses}

\begin{figure}[H]
\centering
\includegraphics[height=5cm]{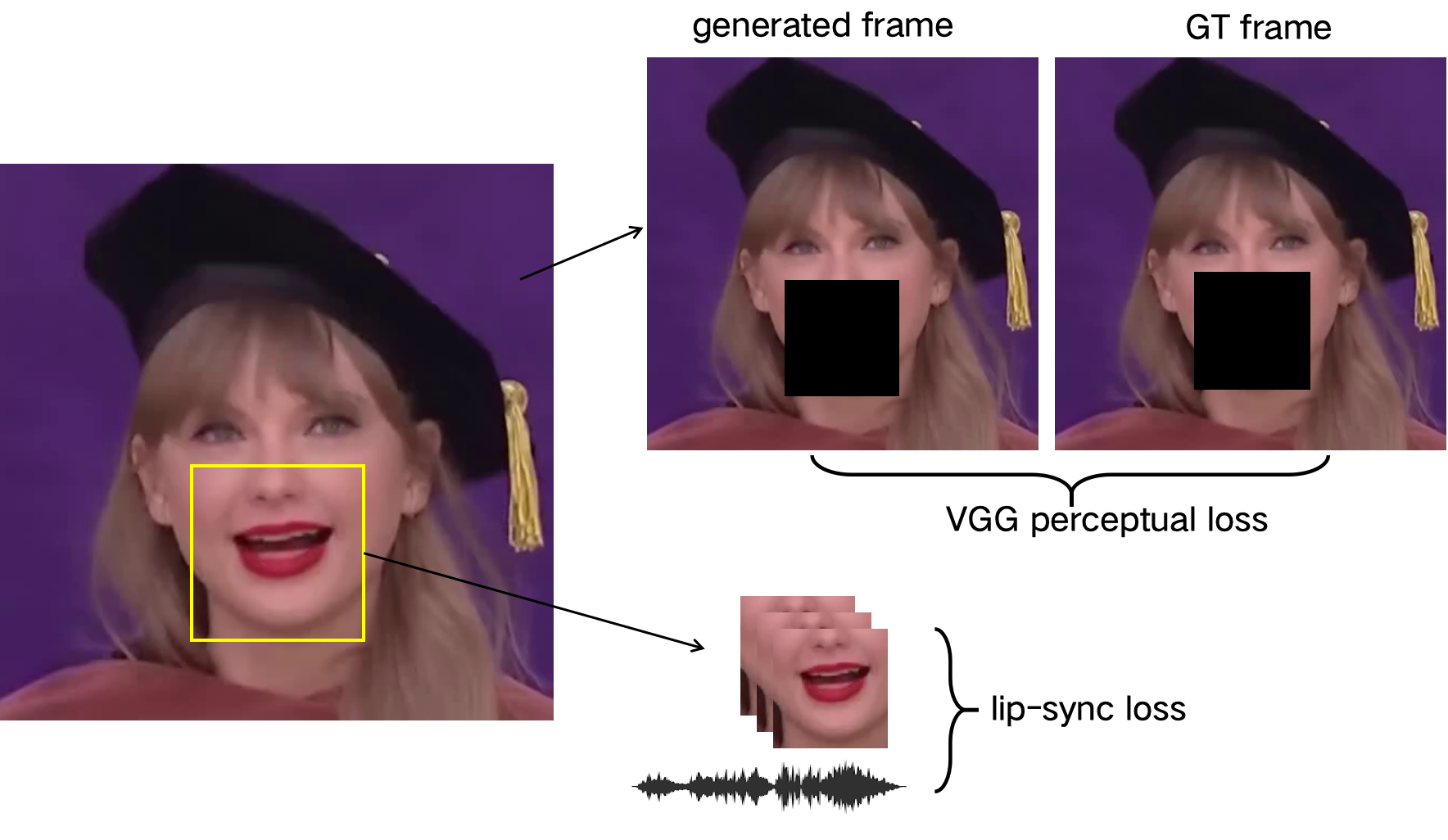}
\caption{\textbf{The combination of two losses.} Perceptual loss and lip-sync loss 
 are used in different areas of the image.}
\label{fig:losses}
\end{figure}

During the training stage, two types of loss functions are employed: perceptual loss~\cite{johnson2016perceptual} and lip-sync loss~\cite{chung2017out,prajwal2020lip}. For different areas of the image, VGG perceptual loss and lip-sync loss are separately calculated as shown in  Fig.~\ref{fig:losses}. The mouth area is related to the driven audio, so the mouth area is cropped for calculating lip-sync loss. During the generation process, the out-of-mouth area is expected to remain unchanged, so we use VGG perceptual loss to minimize the difference between ground truth(GT) and the generated frame.

\noindent\textbf{Perceptual Loss.} We compute perceptual loss in two image scales. Specifically, due to the parameterized expression as mentioned in Sec.~\ref{sec:Audio2exp}, we only change the mouth-related appearance of the generated frame. Therefore, it is possible to compare the source frame and generated frame except for this changed area. As shown in Fig.~\ref{fig:losses},  the image of the mouth area is masked during perceptual loss calculation.
the generate frame $y \in R^{3 \times H \times W}$  and the source frame  $d \in R^{3 \times H \times W}$ are downsampled to $y' \in R^{3 \times \frac{H}{2} \times \frac{W}{2}}$ and $d' \in R^{3 \times \frac{H}{2} \times \frac{W}{2}}$. These paired images \{ $y$, $d$ \} and \{ $y'$, $d'$ \} are encoded by a pre-trained VGG-19 network~\cite{simonyan2014vgg} to compute the perceptual loss.
The $j$th layer of VGG-19 $\phi$ is  $\phi_j$ and total layer number is $M$. The perceptual loss is written as:

\begin{equation}
    \mathcal L_p = \sum_{j=1}^{M} \frac{\left \|\phi_j(d)-\phi_j(y)  \right \|+ \left \|\phi_j(d')-\phi_j(y')  \right \|}{2M}.
\end{equation}

\noindent\textbf{lip-sync Loss.} Following the previous approach Wav2Lip~\cite{prajwal2020lip}, a lip-sync loss is incorporated to enhance the synchronization of lip motion in dubbing videos.  As shown in Fig.~\ref{fig:losses}, we only select the mouth area to improve the sync quality, and the lip-sync network is pre-trained as the same as Wav2Lip~\cite{prajwal2020lip} before model training.
The cosine-similarity loss performs synchronous matching of frames feature $V$ and audio feature $A$. The lip-sync loss is expressed as: 
\begin{equation}
    \mathcal L_{sync} = \frac{V\cdot A}{max\{\left \|V  \right \|_2\cdot\left \| A \right \|_2, \varepsilon \}} .
\end{equation}

The generator minimizes total loss $\mathcal L_{total}$, which is the weighted sum of the perceptual loss and the lip-sync loss.

\begin{equation}
    \mathcal L_{total} = \lambda_p\cdot \mathcal L_p + \lambda_{sync}\cdot \mathcal L_{sync}
\end{equation}

\section{Experiments}

\subsection{Experimental Setup}
\textbf{Implementation Details.} In our experiment, the video sampling rate is 25 FPS and the audio sampling rate is 16KHz. We preprocess all videos by cropping and resizing to $256\times 256$. To synchronize the audio features and the video, We extract the hubert features~\cite{hubertfeature} first. 
We pre-train the audio-video Sync network and face renderer for 3 and 48 hours respectively, and the motion extractor model is a part of the face renderer.
Then we train \textit{Audio2Exp} with the above pre-trained models by a learning rate $1\times 10^{-5} $. And the total training costs 1 day on 8 NVIDIA A10 GPUs. 

\noindent\textbf{Datasets.} We train and evaluate the ControlNet and all the pre-trained models on MEAD~\cite{kaisiyuan2020mead} and HDTF~\cite{hdtf}. MEAD is a high-quality emotional talking-head video set with 8 kinds of emotions. To ensure fair comparisons, we split the MEAD dataset into training and testing sets as official. We download HDTF videos from YouTube with their best resolution and split them into training and testing sets at a ratio of 9:1.

\noindent\textbf{Metrics.} In terms of image generation quality and video synchronization effect, we use SSIM~\cite{SSIM}, Sync~\cite{SYNC}, and Mouth/Face Landmark Distance~\cite{chen2019hierarchical} as metrics respectively. SSIM is used to measure the quality of generated images.
lip-sync evaluates the lip-syn accuracy by calculating the embedding distance between the output video and source audio.
The Mouth/Face Landmarks(M/F-LDM) Distance is used to indicate face consistency by calculating the keypoints between the output image and the ground truth image.

\subsection{Audio-driven Talking Face Generation}

\begin{figure}[H]
\centering
\includegraphics[height=3.5cm]{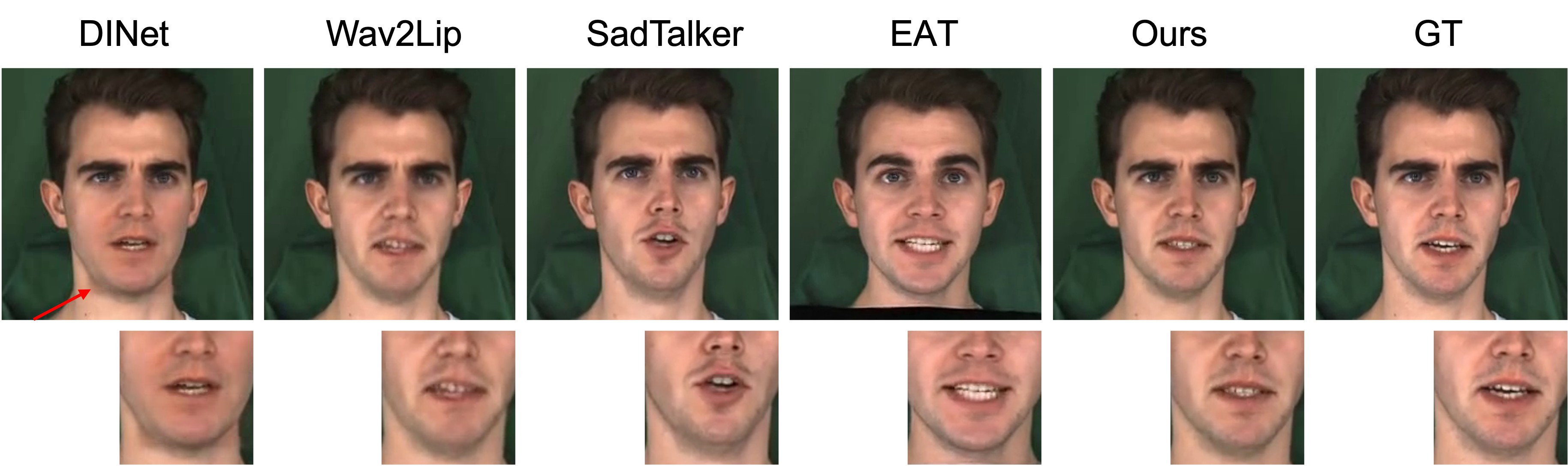}
\caption{Detailed comparisons of different methods. The red arrow points out the mouth box of the DINet.}
\label{fig:detail}
\end{figure}

We conducted the quantitative and qualitative comparisons with the state-of-the-art methods. Both comparisons show our method can generate more accurate mouth shapes and richer facial expressions for lip synchronization. Besides, as shown in Fig.~\ref{fig:detail}, our approach preserves the characteristics of the specific portrait, such as facial appearance and tooth shape. 

\begin{table}[ht]
    \centering
    \small
    \renewcommand{\arraystretch}{1.2}
    \tabcolsep=1.6pt
    \begin{tabular}{c|cccc|cccc}
    
    \hline
        \multicolumn{1}{c}{} & \multicolumn{4}{|c}{HDTF} &\multicolumn{4}{|c}{MEAD} \\
        \cline{2-9} 
        \multicolumn{1}{c|}{}   & SSIM↑  & FID↓ & Sync↓ & M/F-LDM↓  & SSIM↑  & FID↓ & Sync↓ & M/F-LDM↓ \\
    \hline
       Wav2Lip~\cite{prajwal2020lip}   & 0.62 & 41.25 & 0.52 & 2.25/3.27 & \textbf{0.74} & 51.57 & 0.66 & 2.04/2.31\\
       DINet~\cite{zhang2023dinet}     & 0.68 & 32.30 & 0.46 & 1.88/2.78 & 0.73 & \textbf{33.96} & 0.60 & 2.50/2.34\\
       DreamTalk~\cite{ma2023dreamtalk} & 0.60 & 34.07 & 0.50 & 2.72/3.66 & 0.54 & 83.66 & 0.67 &2.97/4.31 \\
       EAT~\cite{gan2023eat}       & 0.68 & 46.99 & 0.49 & 2.23/2.75 & 0.71 & 43.95 & \textbf{0.62} & \textbf{1.85}/\textbf{2.04}\\
       SadTalker~\cite{zhang2023sadtalker} & \textbf{0.69} & \textbf{24.33} & 0.53 & \textbf{1.83}/\textbf{2.56} & 0.64 & 40.92 & 0.64 & 2.75/3.74\\
    \hline
       Ours      & 0.68 & 27.37 & \textbf{0.42} & 2.14/2.93 & 0.71  & 34.62 &  \textbf{0.62}&2.49/2.67 \\
    \hline
    \end{tabular}
    \caption{Comparison with the state-of-the-art methods on HDTF and MEAD dataset. We conduct the comparisons based on same-IDs due to the need for ground truth. SadTalker is evaluated using the fixed pose. Other methods are based on a reference video as a pose sequence.}
    \label{tab:table_metrics}
\end{table}

\begin{figure}[H]
\centering
\includegraphics[height=6.8cm]{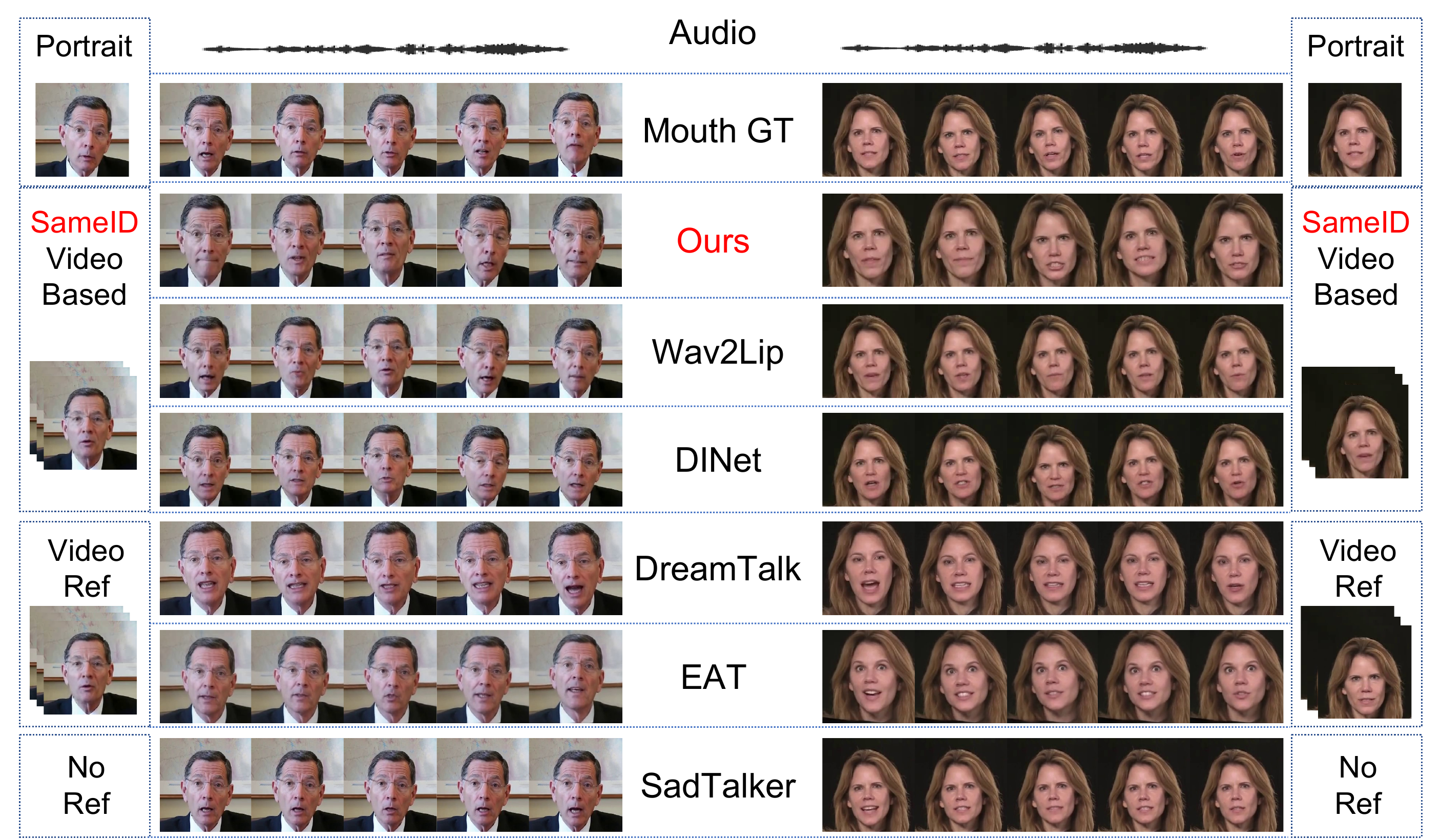}
\caption{Qualitative comparisons with same-ID. The input audio and portrait are the same identity, and all dubbing videos and reference videos come from the same ID.}
\label{fig:sameid}
\end{figure}

\begin{figure}[H]
\centering
\includegraphics[height=6.8cm]{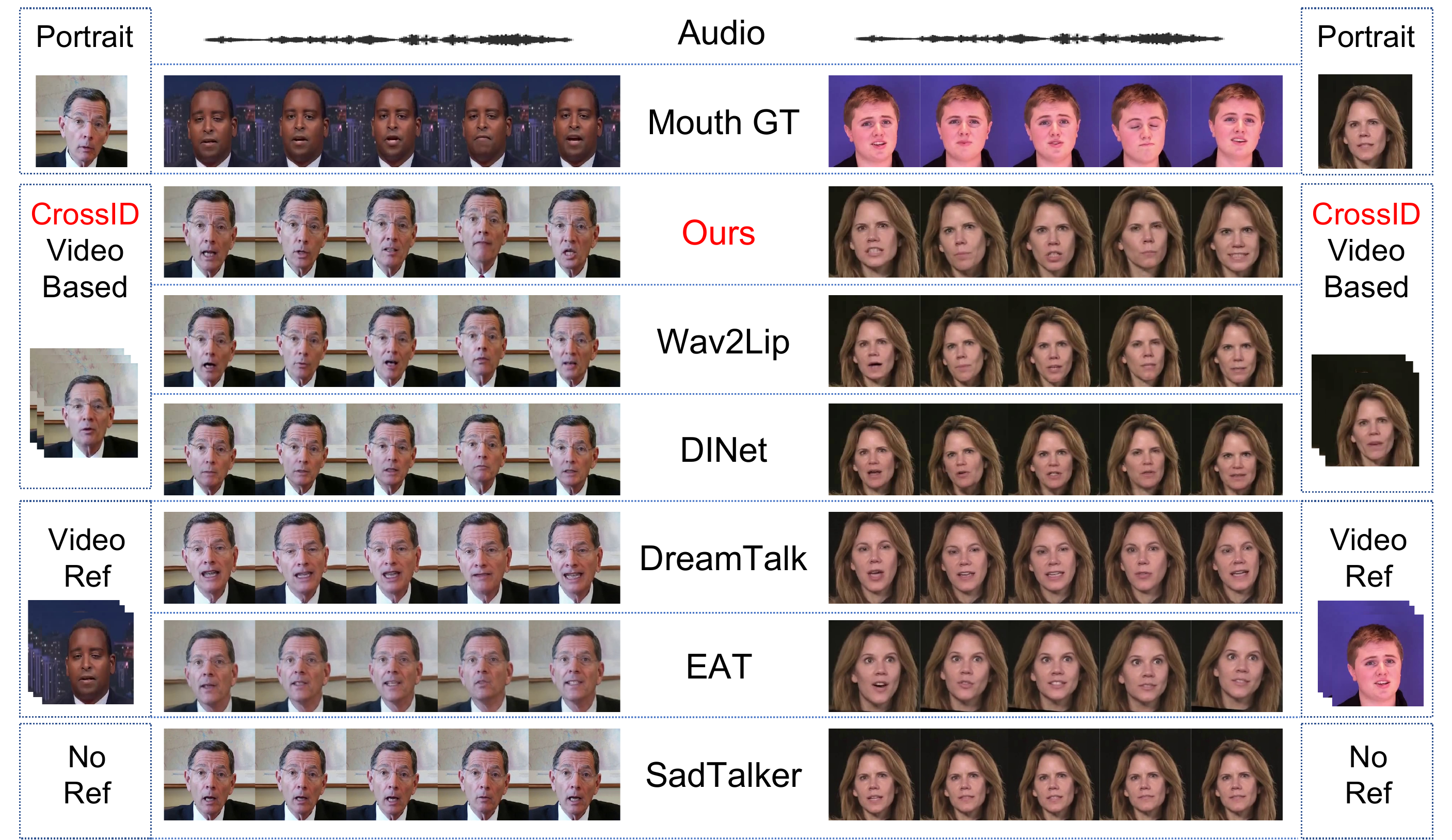}
\caption{Qualitative comparisons with cross-ID. The input audio and portrait are different identities, and the reference video also comes from a different ID.}
\label{fig:crossid}
\end{figure}

\noindent\textbf{Quantitative Comparison.} We conducted the comparisons with Wav2Lip~\cite{prajwal2020lip}, DINet~\cite{zhang2023dinet}, DreamTalk~\cite{ma2023dreamtalk}, EAT~\cite{gan2023eat} and SadTalker~\cite{zhang2023sadtalker}, covering both single image-based and video-based talking face generation methods.
The quantitative analysis comparison is shown in Table.~\ref{tab:table_metrics}. Our method is generally better than previous methods in Sync metric, which indicates the consistency of audio and face image. The other results are highly close overall. 
Compared with traditional GAN-based methods, such as Wav2Lip and DINet, our method is better than them in the image quality field, which is also shown in Fig.~\ref{fig:sameid} and Fig.~\ref{fig:crossid}. The mouth detail of these methods appears blurry.

Furthermore, the other renderer-based methods, DreamTalk and SadTalker, fail to generate the precise lip synchronization face despite using trusted renderers, such as PIRenderer~\cite{ren2021pirenderer}. The superior performance in the Sync metric demonstrates our method’s proficiency in generating lip motion consistent with the reference audio speech.

\noindent\textbf{Qualitative Comparison.} 
We have compared the state-of-the-art methods including video-based dubbing, singe image-based generation with reference video, and singe image-based face animation method, which is shown in Fig.~\ref{fig:sameid} and Fig.~\ref{fig:crossid}.
In order to indicate the impact of different identities and different audio styles, we conducted experiments on the same-ID and cross-ID faces. The ground truth mouth shape is also listed.

It can be seen that our method generates accurate mouth shapes, natural expressions, and good image quality. The mouth of the image generated by Wav2Lip is very blurry and the mouth shape is inaccurate. DINet generates results with a border around the mouth detailed in Fig.~\ref{fig:detail}. The facial expressions of EAT and DreamTalk are relatively uncoordinated, and their mouth shapes are average, which looks like another one. 

The capability of single image-based methods, such as EAT, DreamTalk, and SadTalker, is limited to generating consistent faces, lacking the finesse for realistic and nuanced expressions. For example, no matter what expression the reference video shows, EAT always has wide-open eyes.
Additionally, because of the sequence pose, SadTalker struggles to maintain a consistent head movement for talking face video.
For DreamTalk rows in both same-ID (Fig.~\ref{fig:sameid}) and cross-ID (Fig.~\ref{fig:crossid}), the predicted mouths are exaggerated, and the distorted faces limit the vitality of facial expressions and head movements. Moreover, for different people(left, right),  the opening range of the mouth tends to be a consistent average size. However, our method excels in producing realistic talking faces that not only mirror the specific identity appearance but also achieve precise lip synchronization and superior video quality.
Compared to other methods, our ControlTalk generates more realistic facial expressions and a wider range of head movements based on driven motions.

\subsection{Ablation Study}

\begin{figure}[tb]
\centering
\includegraphics[height=6cm]{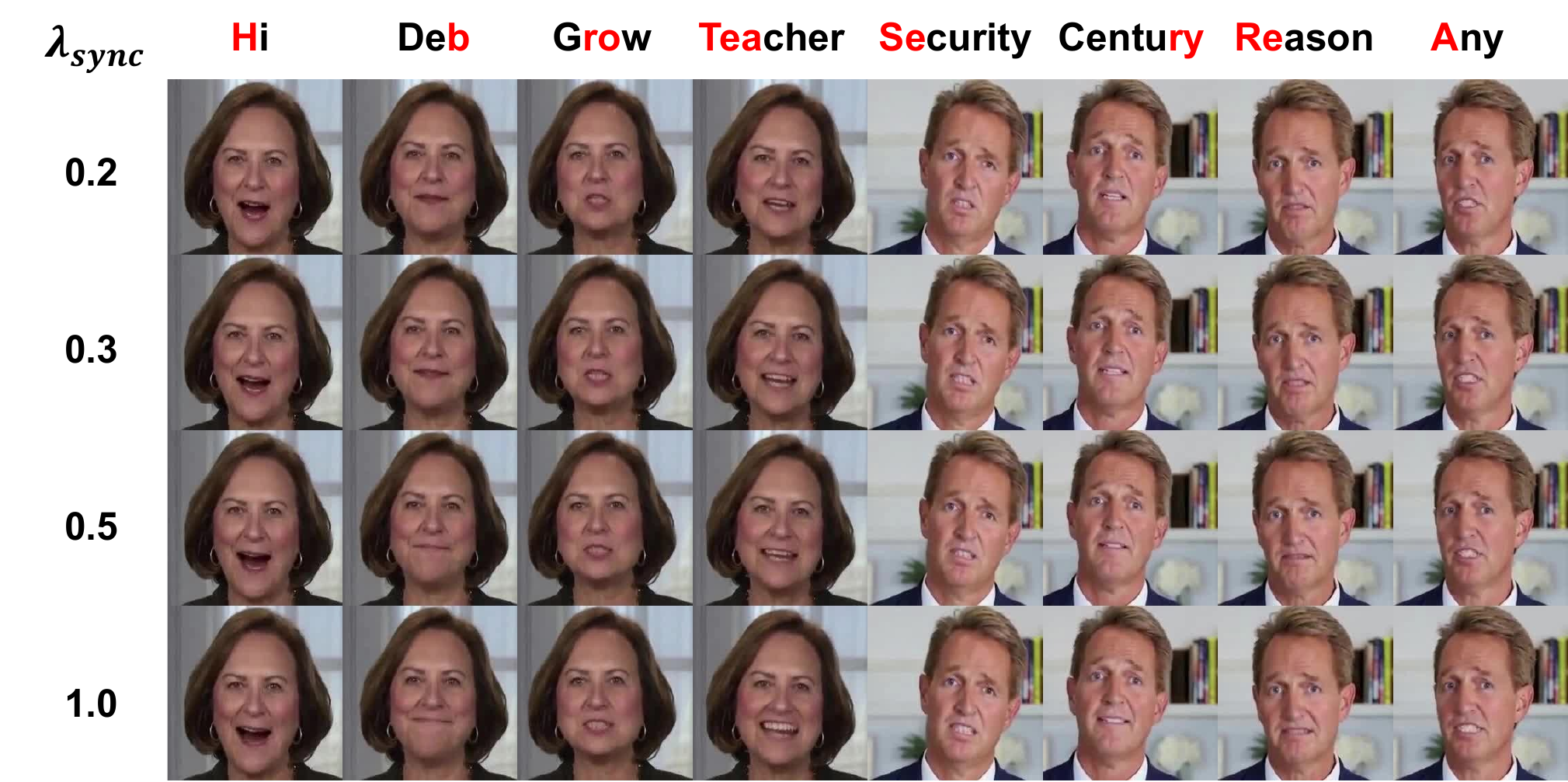}
\caption{Generated frames with different lip-sync ratios. Each frame corresponds to the top syllables in the condition of left lip-sync ratios.}
\label{fig:sync}
\end{figure}

\textbf{Perceptual and Lip-sync Loss.}
The ratio of perceptual and lip-sync loss would affect the accuracy of the lip motion and identity preservation. The greater the weight of lip-sync, the better the mouth shape may be, but the face may be distorted. The greater the weight of perceptual loss, the better the face identity is maintained, but the talking motion may not change significantly. 
We have experimentally verified the ratio of the two losses. The images corresponding to different lip-sync ratios are shown in Fig.~\ref{fig:sync}. 

As the lip-sync ratio increases, we find that changes in the shape of the mouth will extremely match the vocalization,  which leads to unnatural facial expressions.
As shown in the last row of Fig.~\ref{fig:sync}, when the ratio is 1, the mouth shape is already overfitting whenever it is closed or opened. 
Therefore, we use a ratio of 0.3 by default. After experiments, we have found that this is in line with the pronunciation habits of most people.

\begin{figure}[H]
\centering
\includegraphics[height=4.5cm]{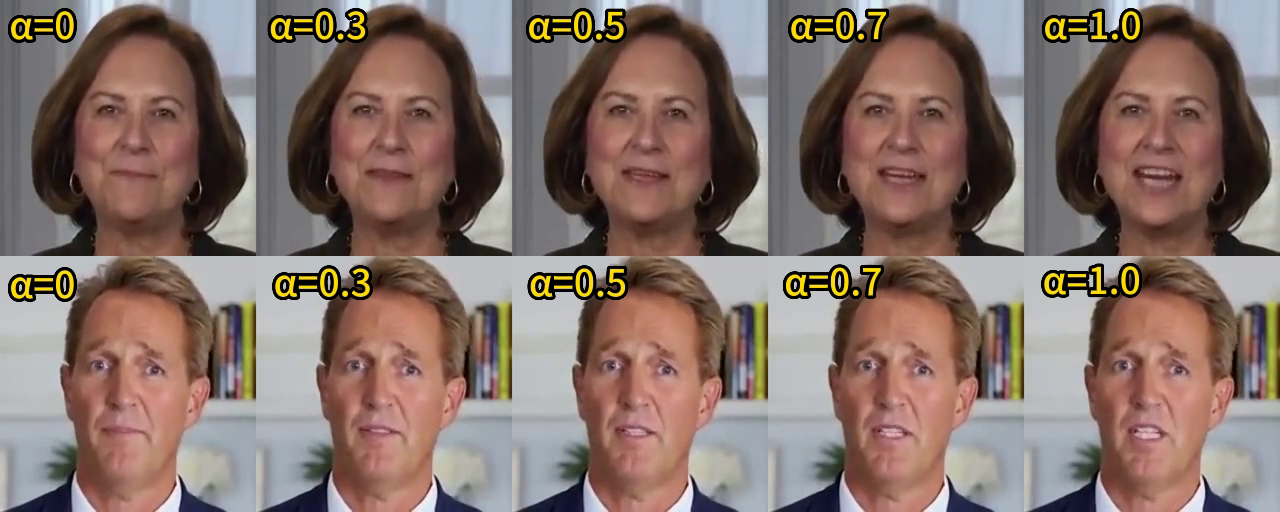}
\caption{Comparison of different $\alpha$. From left to right, the size of the mouth becomes larger as $\alpha$ increases for the same syllable.}
\label{fig:alpha_cmp}
\end{figure}

\noindent\textbf{Mouth Opening Control.} 
\label{sec:exp_mouth_amplitude}
According to the design of the bias coefficient $\alpha$ proposed in Sec.~\ref{sec:mouth_scale}, by adjusting the magnitude of $\alpha$, the degree of mouth opening
is correspondingly affected. The larger the coefficient, the more obvious the mouth shape is. However, a coefficient that is too large may cause the results to be overly exaggerated. If the coefficient is too small, the mouth opening may be too small and the mouth shape may not change significantly. 
Especially, as shown in Fig.~\ref{fig:alpha_cmp}, when $\alpha = 0$, the generated mouth would close whatever the syllable is.
Normally, a value around 0.5 is appropriate, and we can also choose a larger value to achieve a more exaggerated mouth motion.

\subsection{Generalization}
\begin{figure}[H]
\centering
\includegraphics[height=4.5cm]{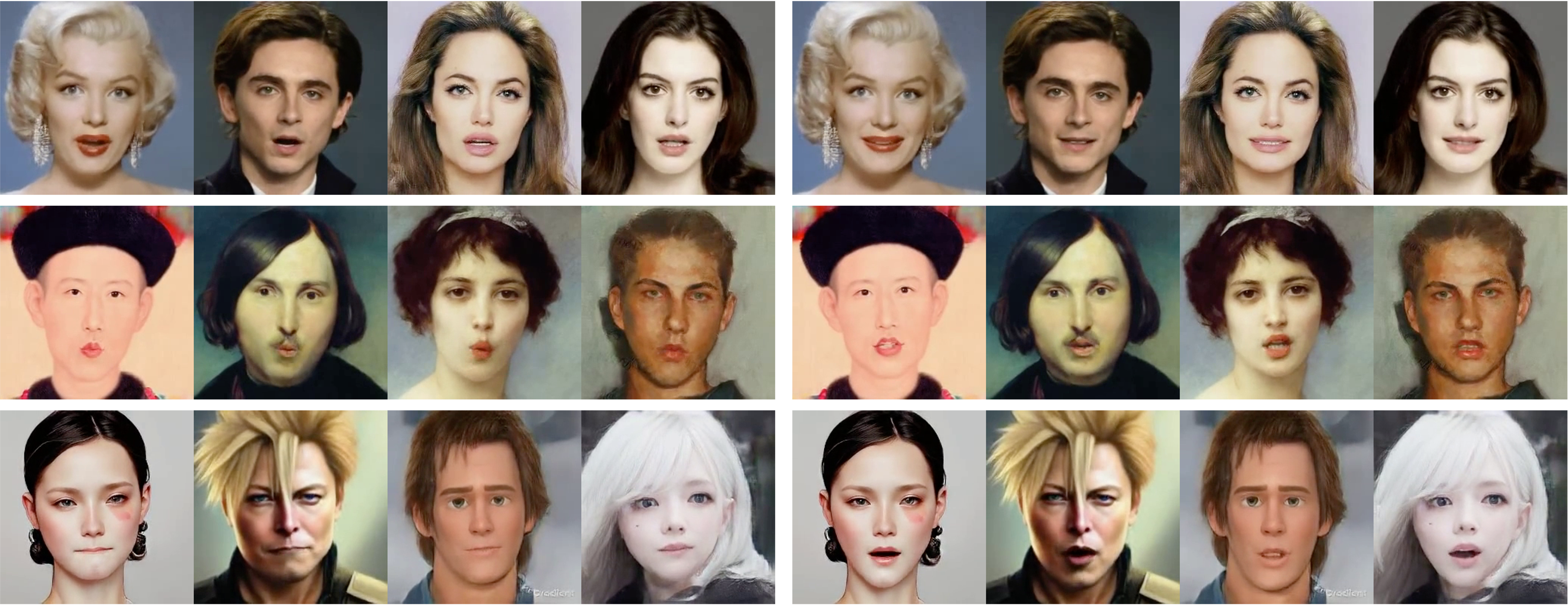}
\caption{Results in different kinds of characters as input portraits. The first row is real humans, the second row is paintings, and the third row is generated images and cartoons.}
\label{fig:cartoon}
\end{figure}

\textbf{Characters Freedom.} Although our model is only trained on real videos, it not only has a good generation of real-human video but is also able to cope with various out-of-domain portraits, even single-image portraits.
The supplemental video demonstrates the capability for different styles of characters, i.e. real humans, paintings, generated faces and cartoons shown in Fig.~\ref{fig:cartoon}.

\begin{figure}[H]
\centering
\includegraphics[height=4cm]{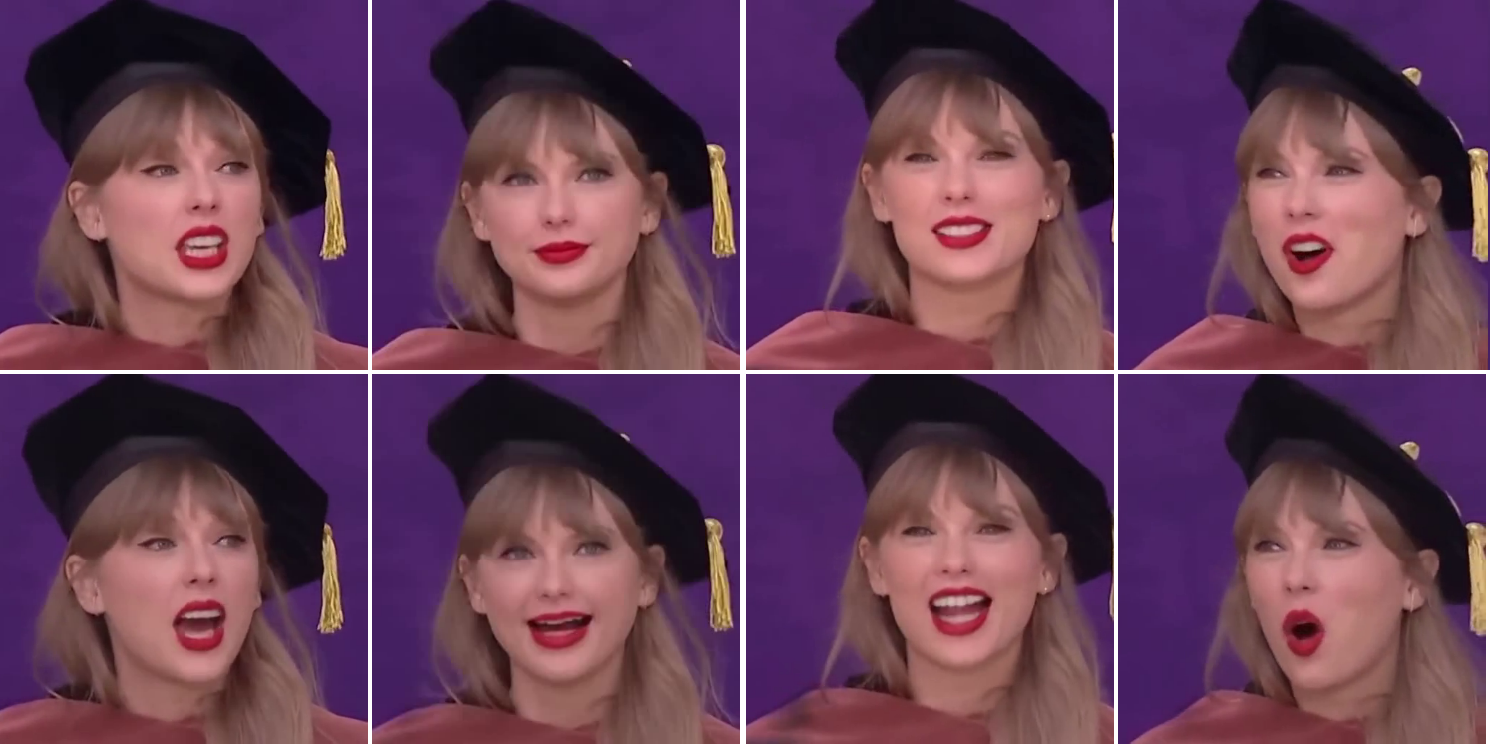}
\caption{Results with French audio. Source video frames are in the first row, and the second row is the generated lip-synchronized frames in French.}
\label{fig:french}
\end{figure}

\noindent\textbf{Languages Expandability.} The language of the training dataset is English without other kinds of languages. However, we test our model over 10 different languages including text-to-speech(TTS) and human speech. For example, a French-driven result is shown in Fig.~\ref{fig:french} with clearly visible changes in the mouth area. More expandability results are shown in our supplemental video.

\begin{figure}[H]
\centering
\includegraphics[height=6cm]{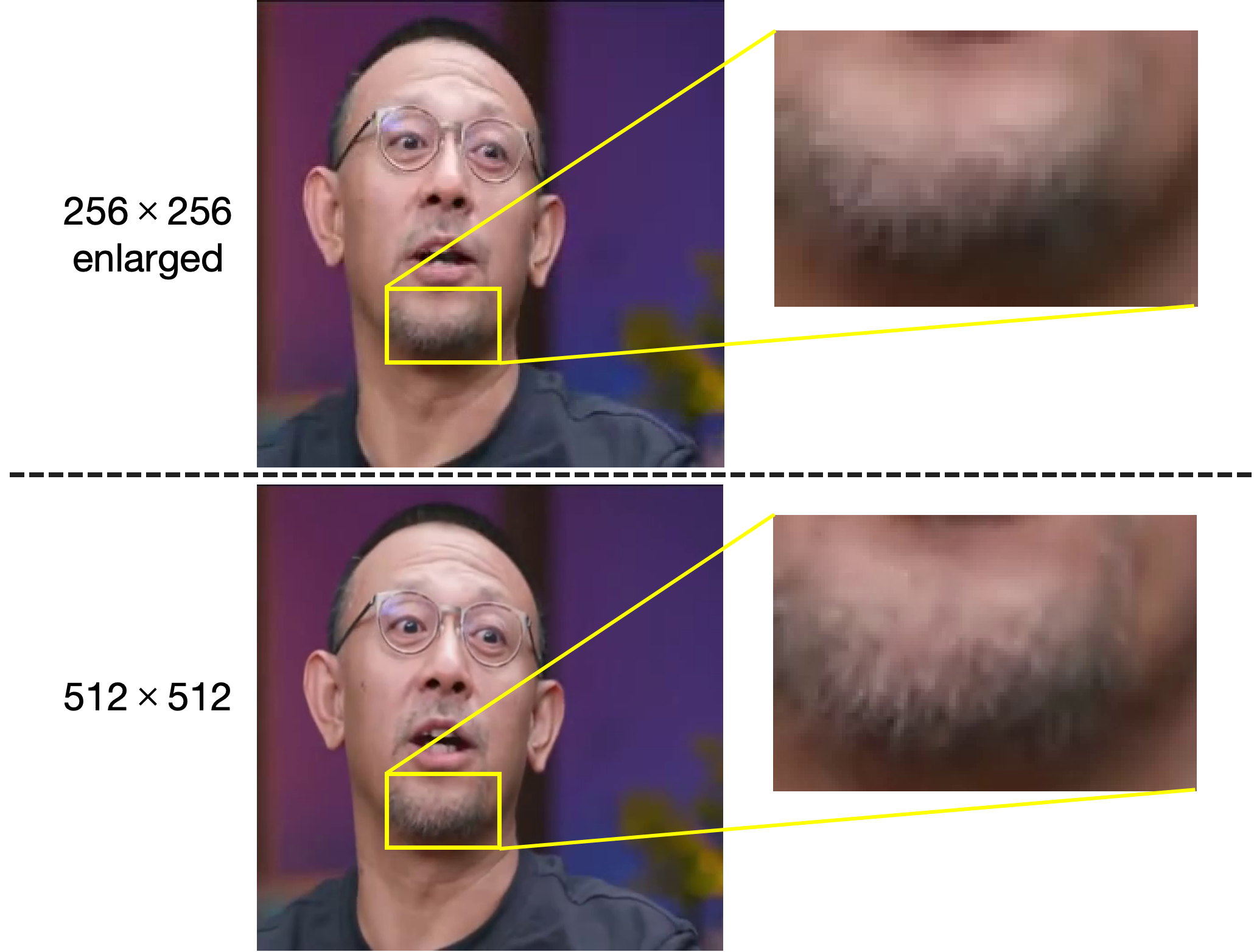}
\caption{Comparisons of $256\times 256$ and $512\times 512$ resolution results.}
\label{fig:512renderer}
\end{figure}

\noindent\textbf{Resolutions Versatility.} The experiments conducted in this paper mainly involve an image resolution of $256\times 256$. However, the {\it Audio2Exp} module is not limited to a single resolution. This is because the {\it Audio2Exp} module manipulates the 3D facial motions based on implicit keypoints.
Therefore, we can generate images of any size as long as the pre-trained renderer supports it.
We have validated the driven effect of the video synthesis model with inputs of $512\times 512$ resolution. 
As shown in Fig.~\ref{fig:512renderer}, when substituting the $512\times 512$ resolution as the input, the generated video/image quality experiences a significant enhancement, with the details of the beard becoming distinctly visible. It demonstrates the independence of our {\it Audio2Exp} module from the face rendering resolution, which can adapt to even higher resolution renderers.

\section{Conclusion and Discussion}

In this paper, we have proposed a novel lip synchronization method ControlTalk, which unifies both image and video-based talking face generation approaches. Our method aims to allow more flexible control while simplifying the generation process. We introduce a lightweight adaptation {\it Audio2Exp} to optimize lip-sync and re-edit the parameterized face expressions. Additionally, the parameterized adaptation allows detailed quantitative control over the mouth-opening shape.
Experiments have proven that our ControlTalk outperforms previous methods in terms of both lip synchronization and video quality, which can be extended to high-resolution video, and can be applied to a diverse range of characters and languages. 


%
%
\bibliographystyle{splncs04}
\bibliography{main}
\end{document}